\documentclass[unnumsec,webpdf,contemporary,large]{oup-authoring-template}%
\usepackage{booktabs}
\usepackage[table]{xcolor}
\graphicspath{{Fig/}}

\theoremstyle{thmstyleone}%
\theoremstyle{thmstyletwo}%
\theoremstyle{thmstylethree}%

\begin{document}

\journaltitle{Journal Title Here}
\DOI{DOI HERE}
\copyrightyear{2022}
\pubyear{2019}
\access{Advance Access Publication Date: Day Month Year}
\appnotes{Paper}

\firstpage{1}


\title[Short Article Title]{Multi-level Self-supervised Pretraining on Compositional Hierarchical Graph for Molecular Property Prediction}

\author[1]{Xiayu Liu \ORCID{0009-0001-5333-0754}}
\author[2]{Zhengyi Lu \ORCID{0009-0003-4723-7187}}
\author[1,$\ast$]{Hou-biao Li\ORCID{0000-0002-7268-307X}}

\authormark{Author Name et al.}

\address[1]{\orgdiv{School of Mathematical Sciences}, \orgname{University of Electronic Science and Technology of China}, \orgaddress{\street{No.2006, Xiyuan Avenue, West Hi-tech Zone}, \postcode{611731}, \state{Chengdu}, \country{China}}}
\address[2]{\orgdiv{Department of Computer Science and Engineering}, \orgname{Oakland University}, \orgaddress{\street{201 Meadow Brook RD, Rochester Hills}, \postcode{48309}, \state{MI}, \country{USA}}}

\corresp[$\ast$]{Corresponding author: Hou-biao Li, \href{email:lihoubiao0189@163.com}{lihoubiao0189@163.com}}

\received{Date}{0}{Year}
\revised{Date}{0}{Year}
\accepted{Date}{0}{Year}



\abstract{Self-supervised pretraining on molecular graphs has emerged as a promising approach for molecular property prediction, yet most existing methods operate at a single structural granularity and treat bond information as auxiliary edge attributes rather than as an independent semantic layer. In this work, we propose MolCHG, a multi-level self-supervised pretraining framework built upon a novel Compositional Hierarchical Graph that organizes molecular structure into four types of nodes across three semantic levels. By introducing a bond graph that operates in parallel with the atom graph, our architecture elevates bond-level information to independently evolving node representations, enabling fragment nodes to aggregate atom-level and bond-level semantics on an equal footing. We design three level-specific pretraining objectives: an atom–bond cross-view contrastive task that aligns the atom-view and bond-view representations within each fragment, a fragment-level functional group prediction task to inject domain-relevant chemical knowledge, and graph-level structure prediction tasks to encode global molecular topology. Experiments on nine MoleculeNet benchmarks demonstrate that MolCHG achieves the best performance on seven datasets across both classification and regression tasks, remaining competitive with the strongest baselines on the rest. Ablation studies further confirm that the multi-level supervision signals are complementary and that each component contributes to the overall performance.}

\keywords{Molecular Property Prediction, Representation Learning, Self-supervised Learning}


\maketitle

\section{Introduction}
Molecular property prediction aims to infer physicochemical properties from molecular structures and plays a central role in drug discovery \cite{xiong2019pushing}, computational biology \cite{wang2026sgac}, and toxicity assessment \cite{tan2023hi}. Because obtaining reliable property labels typically requires expensive and time-consuming wet-lab experiments, large-scale annotated datasets remain scarce, which significantly limits the generalization of purely supervised models. Self-supervised pretraining offers a promising alternative: by learning transferable representations from large volumes of unlabeled molecular data, pretrained models can be adapted to downstream property prediction tasks with considerably less labeled supervision \cite{zhang2024pre,chen2025pretraining,jiang2024dgcl}.  \par

Existing molecular pretraining methods mainly fall into two categories: contrastive learning and predictive learning \cite{zhang2026task}. Contrastive methods improve representation quality by maximizing agreement between augmented views of the same molecule, whereas predictive methods design self-supervised objectives such as recovering masked node attributes or predicting local structural contexts \cite{luo2023improving,li2021effective}. Despite their effectiveness, most existing methods in both families tend to operate at a single structural granularity. For instance, MolCLR \cite{wang2022molecular} defines its contrastive objective at the graph level, while Attribute Masking and Context Prediction \cite{hu2019strategies} focus on node-level objectives that encode local structural and attribute information. Because these objectives are defined at a single level, the learned representations may not fully reflect the hierarchical nature of molecular structures, which spans from individual atoms and bonds through functional substructures to the overall molecular topology.
\par

To move beyond single-granularity pretraining, recent studies
have begun to explore multi-level molecular representation
learning \cite{wu2026hierarchical, li2024molclw, zhang2020motif}.
In particular, fragment-aware methods such as MGSSL
\cite{zhang2021motif} and GraphFP \cite{luong2023fragment} have
shown that chemically meaningful substructures can serve as
effective intermediate semantic units, bridging local atom-level
patterns and global molecular properties. However, these methods
still derive fragment-level semantics primarily from
atom-centered message passing, treating bond information as
auxiliary edge attributes rather than as an independent semantic
layer. In these methods, node representations are iteratively
updated through neighborhood aggregation, whereas edge features
only serve to modulate the messages passed between nodes without
maintaining independently evolving representations. As a result,
even when bond descriptors are provided as edge features, the
bond-level semantics encoded in fragment representations remain
inherently less prominent than atom-level semantics---even
though, from a chemical standpoint, a fragment is jointly
constituted by its atoms and their connecting bonds, both of
which are important compositional elements that deserve explicit representation. Prior supervised studies, including DMPNN
\cite{yang2019analyzing}, CMPNN \cite{song2020communicative},
and DeMol \cite{liu2026enhancing}, have confirmed that
explicitly modeling atom--bond interactions yields more
expressive molecular representations, yet their bond-level
modeling relies on downstream property labels and has not been
extended to the label-free pretraining setting. These
considerations motivate us to decouple bond information from
edge attributes and organize it as an independent structural
layer within a hierarchical pretraining framework.
\par

Among existing hierarchical approaches, HiMol
\cite{zang2023hierarchical} is the most closely related to our
work: it constructs a node--motif--graph hierarchy and designs
self-supervised pretraining tasks over this structure. HimNet
\cite{hong2026hierarchical}, while not a pretraining method,
further demonstrates that hierarchical graph construction
incorporating multi-level interactions can improve molecular
representation quality. However, in both frameworks, bond
information is encoded only as edge attributes and does not
constitute a separate structural layer. We address this limitation by introducing a bond graph that operates in parallel with the atom graph, yielding what we term the Compositional Hierarchical Graph. In this architecture, fragments are explicitly modeled as compositions of atom nodes and bond nodes from two parallel graphs, rather than being derived solely from atom-centered aggregation with bond information as auxiliary edge attributes. On top of this hierarchy, we develop MolCHG, a multi-level self-supervised pretraining framework that assigns a dedicated objective to each structural level: an atom–bond cross-view contrastive objective that aligns atom-view and bond-view representations within each fragment, a fragment-level functional group prediction task to inject domain-relevant semantics, and graph-level structure prediction tasks to encode global molecular topology.
\par

The main contributions of this work are summarized as follows:

\begin{itemize}

\item We propose the Compositional Hierarchical Graph, a heterogeneous molecular graph that organizes atoms, bonds, fragments, and the whole molecule as four co-existing node types across three semantic levels. In this architecture, fragment nodes are explicitly connected to atom nodes and bond nodes from two parallel structural layers, serving as compositional hubs rather than derived summaries of atom-centered pooling. This design elevates bond-level information from auxiliary edge attributes to independently evolving node representations and, more importantly, enables atom-level and bond-level semantics to be aggregated at the fragment level on an equal footing—an interaction that is architecturally absent in prior atom-centric hierarchies.

\item We design three level-specific self-supervised pretraining
tasks---atom-bond cross-view contrastive learning, fragment-level functional group prediction, and graph-level structure prediction---each
targeting a distinct structural granularity of the
Compositional Hierarchical Graph.

\item We conduct extensive experiments on nine MoleculeNet
benchmarks spanning both classification and regression tasks.
MolCHG achieves the best performance on seven datasets while remaining competitive with the strongest baselines on the rest. Ablation studies and representation visualizations further verify the contribution of each proposed component.

\end{itemize}

\section{Methods}\label{sec2}
\subsection{Overview of MolCHG}
\begin{figure*}[t]
\centering
\includegraphics[width=\textwidth]{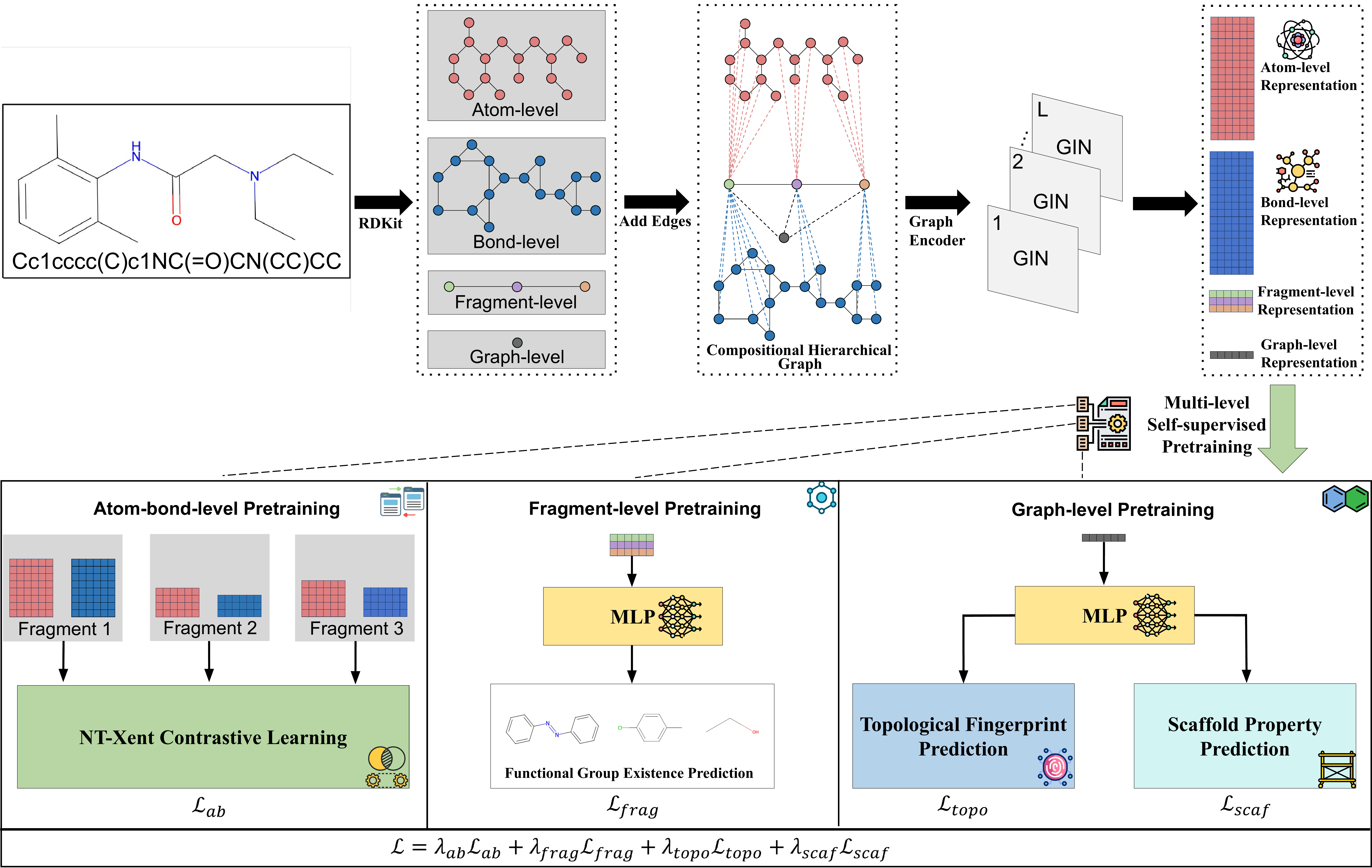}
\caption{Overview of the MolCHG framework.}
\label{fig:Overview}
\end{figure*}

MolCHG is a multi-level self-supervised pretraining framework built upon a novel graph structure that we term the Compositional Hierarchical Graph. The overall architecture consists of two main components: (i) the construction of the Compositional Hierarchical Graph, which organizes molecular structure into multiple semantic levels, and (ii) a set of level-specific pretraining objectives that operate over this hierarchy. An illustration of the overall framework is provided in Figure ~\ref{fig:Overview}.

\par

Given a molecular input, we first construct the Compositional Hierarchical Graph, which comprises four types of nodes arranged across three structural levels. At the bottom level, an atom graph and a bond graph operate in parallel: atom nodes represent individual atoms and bond nodes represent individual bonds, with edges in both graphs encoding only topological connectivity without carrying additional feature information. At the intermediate level, fragment nodes are introduced by decomposing the molecule into chemically meaningful substructures. Each fragment node is connected to its constituent atom nodes and bond nodes, serving as a compositional hub that aggregates both atomic and bond-level semantics. In addition, fragment nodes are interconnected according to the connectivity relationships among their corresponding substructures, forming a fragment-level graph that captures the higher-order topology of the molecule. At the top level, a virtual graph node, whose representation is iteratively refined through message passing, is connected to all fragment nodes.
\par

On top of this hierarchical structure, we design three self-supervised pretraining objectives, each targeting a specific level of the hierarchy. At the atom-bond level, we employ a cross-view contrastive objective that maximizes the agreement between atom-view and bond-view representations within each fragment, encouraging atom-level and bond-level semantics that jointly constitute the same fragment to align in a shared representation space. At the fragment level, we formulate a functional group prediction task that injects chemically meaningful supervision into the intermediate layer. At the graph level, we introduce structure prediction tasks based on molecular scaffolds and topological fingerprints to capture the global structural organization. The three objectives are jointly optimized to provide complementary supervision signals ranging from local compositional patterns to global molecular properties. 
\par

\subsection{Compositional Hierarchical Graph Construction}

Conventional molecular graph representations encode bond information as edge attributes that modulate the messages passed between atom nodes during neighborhood aggregation. While this design retains bond descriptors within the computation, it does not grant bonds independently evolving representations, leaving the resulting representations inherently atom-centric. To address this limitation, the Compositional Hierarchical Graph organizes bonds as an independent node layer that operates in parallel with the atom graph, and introduces fragment nodes as compositional intermediaries that aggregate semantics from both layers. The remainder of this section details its construction.
\par

Given a SMILES string, we first use RDKit to construct the atom graph $\mathcal{G}_a = (\mathcal{V}_a, \mathcal{E}_a)$, where each node $v\in \mathcal{V}_{a}$ represents an atom and each edge $(v_{i},v_{j})\in \mathcal{E}_a$ indicates the existence of a chemical bond between atom $v_{i}$ and $v_{j}$. We simultaneously construct a bond graph $\mathcal{G}_b=(\mathcal{V}_b,\mathcal{E}_b)$, in which each node $u\in \mathcal{V}_b$ corresponds to a bond in the original molecule and two bond nodes are connected by an edge if their corresponding bonds share a common atom. For example, consider three atoms connected as $A$--$B$--$C$ with bonds $e_{1}(A,B)$ and $e_{2}(B,C)$. In the bond graph, $e_{1}$ and $e_{2}$ become two nodes connected by an edge because they share atom $B$.
\par 

To introduce the intermediate structural level, we decompose each molecule into chemically meaningful substructures using the Principal Subgraph Mining algorithm \cite{kong2022molecule}. Unlike rule-based fragmentation methods, such as BRICS \cite{degen2008art}, which tend to produce a large and diverse set of fragment types, Principal Subgraph Mining identifies a compact vocabulary of recurring substructures from the pretraining corpus. A smaller vocabulary concentrates the training signal on the frequently occurring structural patterns, allowing the model to learn more robust fragment-level representations during pretraining. Each fragment is represented by a fragment node $f\in \mathcal{V}_f$. We deliberately avoid initializing fragment nodes via pooling over their constituent atom nodes, as is done in several prior methods \cite{panapitiya2026fragnet}. This design choice preserves the role of fragment nodes as intermediary hubs: on the one hand, their initial features provide connected atom and bond nodes with contextual information about the local substructure they belong to; on the other hand, their representations are further informed by atom-level and bond-level information through message passing, allowing atom and bond semantics to interact through them.
\par

\begin{table*}[t]
\centering
\caption{Initial node features for each node type in the Compositional Hierarchical Graph. Features are designed to exclude information targeted by the pretraining objectives.}
\label{tab:node_features}
\begin{tabular}{p{2.1cm}p{13.8cm}c}
\toprule
\textbf{Node Type} & \textbf{Description} & \textbf{Dim} \\
\midrule
Atom Node & Atomic number, degree, formal charge, radical electrons, hybridization (one-hot), scaled atomic mass, total hydrogens, chirality center indicator, chirality type & 15 \\
Bond Node & Bond existence indicator, bond type (single, double, triple), whether connecting different elements, bond direction, formal charge difference between endpoints, conjugation indicator, stereo chemistry (one-hot) & 15 \\
Fragment Node & Number of atoms, bonds, heteroatoms, C/N/O atoms, halogens, single/double/triple bonds, average atomic mass, average degree, sum of formal charges, total hydrogens, total valence & 15 \\
Graph Node & Number of atoms, bonds, fragments, average/max/min fragment atom count, average/max/min fragment bond count, small/medium/large/single-atom fragment counts, atom and bond count variance & 15 \\
\bottomrule
\end{tabular}
\end{table*}

Each fragment node is connected to all of its constituent atom nodes in $\mathcal{G}_{a}$ and to all of its constituent bond nodes in $\mathcal{G}_{b}$. Two fragment nodes are connected by an edge if their corresponding substructures are adjacent in the original molecule, forming a fragment-level graph $\mathcal{G}_f=(\mathcal{V}_f, \mathcal{E}_f)$ that captures the higher-order topology. Finally, a virtual graph node $g$ is introduced and connected to all fragment nodes, serving as the graph-level representation hub. The initial features for all four node types are designed to capture relevant chemical or structural descriptors while deliberately excluding any information targeted by the pretraining objectives, thereby preventing leakage. The initial features for each node type are listed in Table \ref{tab:node_features}.
\par 

The resulting Compositional Hierarchical Graph is defined as $\mathcal{G}=(\mathcal{V},\mathcal{E})$, where $\mathcal{V}=\mathcal{V}_a \cup \mathcal{V}_b\cup \mathcal{V}_{f}\cup \{g\}$ denotes the union of all node sets and $\mathcal{E}=\mathcal{E}_a \cup \mathcal{E}_b\cup \mathcal{E}_f \cup \mathcal{E}_{af} \cup \mathcal{E}_{bf} \cup \mathcal{E}_{fg}$ denotes the union of intra-level edges and inter-level edges that connect atom nodes to fragment nodes $\mathcal{E}_{af}$, bond nodes to fragment nodes $\mathcal{E}_{bf}$, and fragment nodes to virtual graph node $\mathcal{E}_{fg}$. Throughout the entire graph, all edges serve solely as topological connectors and carry no additional feature information. Because fragment nodes are explicitly connected to both atom nodes and bond nodes, they serve as compositional intermediaries through which atom-level and bond-level semantics can interact during message passing---an interaction that is architecturally absent in conventional atom-centric hierarchies where bond information resides only in edge attributes.

\subsection{Multi-level Pretraining Objectives}
The Compositional Hierarchical Graph organizes molecular structure into three semantic levels, each capturing a distinct aspect of molecular composition. We design a dedicated self-supervised objective for each level, so that the resulting supervision signals span from local compositional patterns to global structural properties.
\par 

At the atom-bond level, we leverage the dual connectivity between fragment nodes and their constituent atom and bond nodes to define a cross-view contrastive objective that aligns the atom-view and bond-view representations of each fragment in a shared semantic space. For a given fragment $f_k$, we apply mean pooling over its constituent atom node representations and bond node representations separately, yielding two summary vectors. Each summary vector is then passed through a learnable projection head, producing $\tilde{z}_{a}^{(k)}$ and $\tilde{z}_{b}^{(k)}$ for the atom view and bond view, respectively. Within a mini-batch containing $N$ valid fragments (excluding single-atom fragments), the atom summary and bond summary of the same fragment form a positive pair, while all cross-fragment pairings serve as negatives. The contrastive loss follows the symmetric NT-Xent formulation:
\begin{equation}
    \ell_{a \to b}^{(k)} = -\log \frac{\exp \left(\mathrm{sim}\left(\tilde{z}_a^{(k)}, \tilde{z}_b^{(k)}\right)/\tau\right)}{\sum_{j=1}^{N}
\exp\left(\mathrm{sim}\left(\tilde{z}_a^{(k)}, \tilde{z}_b^{(j)}\right)/\tau\right)}
\end{equation}
where $\mathrm{sim}(\cdot,\cdot)$ denotes cosine similarity and $\tau$ is a temperature hyperparameter. The reverse direction $\ell_{b\to a}^{(k)}$ is defined analogously. The overall atom-bond cross-view contrastive loss is:
\begin{equation}
    \mathcal{L}_{ab}=\frac{1}{2N}\sum_{k=1}^{N}(\ell_{a \to b}^{(k)}+\ell_{b \to a}^{(k)})
\end{equation}
This objective aligns the atom-view and bond-view representations of each fragment in a shared semantic space, reinforcing the semantic agreement between atom-level and bond-level information that jointly constitutes the same fragment. Simultaneously, the negative-sample mechanism pushes representations of different fragments apart, enhancing their discriminability. Notably, fragments of the same substructure type appearing in different molecules are also treated as negative pairs, since the same substructure may serve different functional roles across distinct molecular contexts, and the information propagated to fragment nodes through message passing varies with the surrounding molecular environment.
\par

At the fragment level, we inject chemically meaningful supervision into the intermediate layer of the hierarchy by formulating a multi-label classification task. Each fragment node is trained to predict which functional groups are present within the corresponding substructure. We select the $C$ most frequently occurring functional groups from the pretraining corpus as the target label set, and use RDKit substructure matching to produce a binary label vector $y_{f}^{(k)}\in \{0,1\}^{C}$ for each fragment $f_k$. The fragment node representation after message passing is fed into a prediction head to produce logits $\tilde{y}_f^{(k)}\in R^{C}$, and the loss is computed via binary cross-entropy:
\begin{equation}
    \mathcal{L}_{frag}=-\frac{1}{|\mathcal{F}|}\sum_{f_k\in \mathcal{F}} \sum_{c=1}^{C}[y_{f,c}^{(k)}\log \sigma(\tilde{y}_{f,c}^{(k)})+(1-{y}_{f,c}^{(k)})\log(1-\sigma(\tilde{y}_{f,c}^{(k)}))]
\end{equation}
where $\sigma(\cdot)$ is the sigmoid function and $\mathcal{F}$ denotes the set of valid fragments. By training fragment nodes to recognize chemically meaningful substructure types, this objective encourages the intermediate representations to encode domain-relevant semantic knowledge that complements the compositional information captured at the atom-bond level.
\par 

At the graph level, the virtual graph node participates in the message passing process to form a unified molecular representation. We design two complementary prediction tasks at this level. The first targets topological fingerprints: for each molecule, we compute a $D$-bit topological fingerprint using RDKit, which captures the global topological connectivity of the molecule. The graph node representation is passed through a prediction head, and the loss is computed via binary cross-entropy:
\begin{equation}
    \mathcal{L}_{topo}=-\frac{1}{D}\sum_{d=1}^{D}[y_d \log\sigma(\tilde{y}_d)+(1-y_{d})\log(1-\sigma(\tilde{y}_d))]
\end{equation}
The second task focuses on the molecular scaffolds, which characterize the core ring systems of a molecule. We predict five scaffold-related properties: the total number of rings, the number of aromatic rings, and three binary indicators denoting the presence of fused rings, heterocyclic rings, and bridged rings. These properties collectively characterize the global topological organization of the molecular scaffold. The scaffold loss is defined as:
\begin{equation}
\mathcal{L}_{scaf}=\mathcal{L}_{ring}+\mathcal{L}_{aro}+\mathcal{L}_{bin}
\end{equation}
where $\mathcal{L}_{ring}$ and $\mathcal{L}_{aro}$ are cross-entropy losses for the total ring count and aromatic ring count classification, respectively, and $\mathcal{L}_{bin}$ is a binary cross-entropy loss over the three structural indicators.
\par 

The four loss terms across three structural levels are jointly optimized during pretraining:
\begin{equation}  \mathcal{L}=\lambda_{ab}\mathcal{L}_{ab}+\lambda_{frag}\mathcal{L}_{frag}+\lambda_{topo}\mathcal{L}_{topo}+\lambda_{scaf}\mathcal{L}_{scaf}
\end{equation}
where $\lambda_{ab}$, $\lambda_{frag}$, $\lambda_{topo}$, and $\lambda_{scaf}$ are hyperparameters controlling the relative 
contribution of each task.

\section{Experiments}\label{sec3}
\begin{table*}[t]
\centering
\scriptsize
\renewcommand{\arraystretch}{1.05}
\caption{Performance comparison on molecular property prediction benchmarks. Classification tasks are evaluated by ROC-AUC (\%, $\uparrow$) and regression tasks by RMSE ($\downarrow$). The best results are highlighted in \textbf{bold}.}
\label{tab:main_results}
\setlength{\tabcolsep}{3pt}
\begin{tabular}{l*{6}{c}|*{3}{c}}
\toprule
 & \multicolumn{6}{c|}{Classification (ROC-AUC $\uparrow$)} & \multicolumn{3}{c}{Regression (RMSE $\downarrow$)} \\
 \cmidrule(lr){1-7} \cmidrule(lr){8-10}
Methods & BACE & BBBP & ClinTox & SIDER & Tox21 & HIV & ESOL & FreeSolv & Lipophilicity \\
\midrule
ContextPred & $78.39{\pm}0.58$ & $69.10{\pm}0.29$ & $55.63{\pm}1.35$ & $61.83{\pm}0.60$ & $73.26{\pm}0.59$ & $72.04{\pm}0.48$ & $2.190{\pm}0.026$ & $3.195{\pm}0.058$ & $1.053{\pm}0.048$ \\
AttrMasking & $75.95{\pm}0.50$ & $67.12{\pm}0.45$ & $60.11{\pm}1.19$ & $61.21{\pm}0.65$ & $73.37{\pm}0.55$ & $72.71{\pm}0.70$ & $2.954{\pm}0.087$ & $4.023{\pm}0.039$ & $0.982{\pm}0.052$ \\
EdgePred    & $74.29{\pm}1.37$ & $64.73{\pm}1.10$ & $61.62{\pm}1.25$ & $60.18{\pm}0.76$ & $70.32{\pm}1.62$ & $70.55{\pm}1.68$ & $2.368{\pm}0.070$ & $3.192{\pm}0.023$ & $1.085{\pm}0.061$ \\
Infomax     & $77.80{\pm}0.46$ & $68.39{\pm}0.64$ & $58.62{\pm}0.83$ & $59.02{\pm}0.56$ & $72.66{\pm}0.16$ & $73.55{\pm}0.47$ & $2.953{\pm}0.049$ & $3.033{\pm}0.026$ & $0.970{\pm}0.023$ \\
JOAO        & $74.94{\pm}1.35$ & $71.63{\pm}1.11$ & $77.02{\pm}1.64$ & $63.55{\pm}0.81$ & $73.67{\pm}1.06$ & $77.55{\pm}1.94$ & $1.978{\pm}0.029$ & $3.282{\pm}0.002$ & $1.093{\pm}0.097$ \\
JOAOv2      & $74.38{\pm}1.71$ & $71.98{\pm}0.18$ & $65.22{\pm}0.75$ & $59.88{\pm}1.72$ & $73.95{\pm}1.88$ & $77.13{\pm}1.51$ & $2.144{\pm}0.009$ & $3.842{\pm}0.012$ & $1.116{\pm}0.024$ \\
GraphCL     & $77.80{\pm}0.46$ & $68.39{\pm}0.64$ & $61.62{\pm}1.25$ & $61.83{\pm}0.60$ & $73.26{\pm}0.59$ & $73.55{\pm}0.47$ & $1.390{\pm}0.363$ & $3.166{\pm}0.027$ & $1.014{\pm}0.018$ \\
GraphLoG    & $76.60{\pm}1.04$ & $66.75{\pm}0.32$ & $53.76{\pm}0.95$ & $59.09{\pm}0.53$ & $71.64{\pm}0.49$ & $73.76{\pm}0.29$ & $1.542{\pm}0.026$ & $2.335{\pm}0.052$ & $0.932{\pm}0.052$ \\
GraphFP     & $80.28{\pm}3.06$ & $72.05{\pm}1.17$ & $76.80{\pm}1.83$ & \boldmath{$65.93{\pm}3.09$} & $77.35{\pm}1.40$ & $75.71{\pm}1.39$ & $2.136{\pm}0.096$ & $2.528{\pm}0.016$ & $1.371{\pm}0.058$ \\
MICRO-Graph & $63.57{\pm}1.55$ & $67.21{\pm}1.85$ & $77.56{\pm}1.56$ & $60.34{\pm}0.96$ & $71.79{\pm}1.70$ & $76.73{\pm}1.07$ & $0.842{\pm}0.055$ & $1.865{\pm}0.061$ & $0.851{\pm}0.073$ \\
MGSSL       & $82.03{\pm}3.79$ & $79.52{\pm}1.98$ & $75.84{\pm}1.82$ & $57.46{\pm}1.45$ & $74.82{\pm}1.60$ & $77.45{\pm}2.94$ & $2.936{\pm}0.071$ & $2.940{\pm}0.051$ & $1.106{\pm}0.077$ \\
GROVER      & $81.13{\pm}0.14$ & $87.15{\pm}0.06$ & $72.53{\pm}0.14$ & $57.53{\pm}0.23$ & $68.59{\pm}0.24$ & $75.04{\pm}0.13$ & $1.237{\pm}0.403$ & $2.712{\pm}0.327$ & $0.823{\pm}0.027$ \\
SimSGT      & $79.75{\pm}1.28$ & $71.51{\pm}1.75$ & $74.11{\pm}1.05$ & $59.74{\pm}1.32$ & $76.23{\pm}1.27$ & $78.13{\pm}1.07$ & $0.932{\pm}0.026$ & $1.953{\pm}0.038$ & $0.771{\pm}0.041$ \\
MoAMa       & $81.32{\pm}1.06$ & $85.89{\pm}0.61$ & $77.11{\pm}1.67$ & $62.69{\pm}0.37$ & $78.29{\pm}0.55$ & $78.11{\pm}0.64$ & $1.125{\pm}0.029$ & $2.072{\pm}0.053$ & $1.085{\pm}0.024$ \\
HiMol       & $86.14{\pm}1.59$ & $82.24{\pm}1.77$ & $77.47{\pm}1.71$ & $61.82{\pm}2.02$ & $80.16{\pm}1.04$ & $76.95{\pm}1.20$ & $0.971{\pm}0.047$ & $1.653{\pm}0.085$ & $0.826{\pm}0.04$ \\
S-CGIB      & $86.46{\pm}0.81$ & $88.75{\pm}0.49$ & $78.58{\pm}2.01$ & $64.03{\pm}1.04$ & $80.94{\pm}0.17$ & \boldmath{$78.33{\pm}1.34$} & $0.816{\pm}0.019$ & $1.648{\pm}0.074$ & $0.762{\pm}0.042$ \\
\midrule
\rowcolor{gray!15}
MolCHG (Ours) & \boldmath{$87.20{\pm}0.86$} & \boldmath{$89.46{\pm}0.54$} & \boldmath{$80.83{\pm}1.69$} & $63.89{\pm}1.20$ & \boldmath{$82.51{\pm}1.20$} & $77.83{\pm}0.67$ & \boldmath{$0.797{\pm}0.027$} & \boldmath{$1.523{\pm}0.062$} & \boldmath{$0.745{\pm}0.036$} \\
\bottomrule
\end{tabular}
\end{table*}

\subsection{Experimental Setup}
We evaluate MolCHG on molecular property prediction tasks using benchmarks from MoleculeNet \cite{wu2018moleculenet}. The downstream datasets include six classification tasks (BACE, BBBP, ClinTox, SIDER, Tox21, and HIV) evaluated by ROC-AUC (\%), and three 
regression tasks (ESOL, FreeSolv, and Lipophilicity) evaluated by RMSE. Following S-CGIB \cite{lee2025pre}, all downstream datasets are randomly split into training, validation, and test sets with a ratio of 6:2:2. We report the mean and standard deviation over 10 independent runs.
\par

For pretraining, we use the same subset of 250K unlabeled molecules from the ZINC15 dataset \cite{sterling2015zinc} as used by HiMol to ensure a fair comparison. The GNN encoder is a 5-layer GIN \cite{xu2018powerful} with a hidden dimension of 300, dropout rate of 0.5, and sum-based jumping knowledge aggregation. The fragment vocabulary is constructed using the Principal Subgraph Mining algorithm with a vocabulary size of 800. The model is pretrained for 100 epochs using the Adam optimizer with a learning rate of $1\mathrm{e}{-3}$ and weight decay of $1\mathrm{e}{-5}$. The batch size is set to 256 to provide sufficient negative samples for the atom-bond cross-view contrastive objective, with the temperature $\tau$ set to 0.1. The loss weights are set to $\lambda_{ab}=0.2, \lambda_{frag}=0.4, \lambda_{topo}=0.4$ and $\lambda_{scaf}=0.4$, where a smaller weight is assigned to the contrastive loss since its magnitude is typically larger than that of the predictive losses. For finetuning, the pretrained GNN encoder weights are transferred to the downstream model. To fully exploit the multi-level representations learned during pretraining, we apply mean pooling over atom nodes, bond nodes, and fragment nodes within each molecule separately, and concatenate the three pooled vectors with the graph node representation to form the final molecular representation. A single linear layer is then appended as the prediction head, mapping the concatenated representation to the target property. The model is finetuned for 100 epochs using the Adam optimizer with a learning rate of $1\mathrm{e}{-3}$ and a batch size of 32.

\subsection{Baseline}
We compare MolCHG against a comprehensive set of self-supervised pretraining baselines, which can be grouped into three categories. The first category is node-level pretraining methods, including ContextPred \cite{hu2019strategies}, AttrMasking \cite{hu2019strategies}, and EdgePred \cite{hamilton2017inductive}, which design pretraining objectives at the node level such as context prediction, attribute recovery, and edge prediction. The second category is contrastive learning methods, including Infomax \cite{velivckovic2018deep}, JOAO \cite{you2021graph}, JOAOv2 \cite{you2021graph}, GraphCL \cite{you2020graph}, and GraphLoG \cite{xu2021self}, which learn molecular representations by contrasting augmented or multi-view graph representations. The third category is subgraph/fragment-level pretraining methods, including GraphFP \cite{luong2023fragment}, MICRO-Graph \cite{zhang2020motif}, MGSSL \cite{zhang2021motif}, GROVER \cite{rong2020self}, SimSGT \cite{liu2023rethinking}, MoAMa \cite{inae2023motif}, and HiMol \cite{zang2023hierarchical}, which leverage substructure or fragment information to capture higher-order structural patterns during pretraining. We also compare with S-CGIB \cite{lee2025pre}, which compresses molecular graphs into core subgraphs through a graph information bottleneck principle. The results of all baseline methods except HiMol are taken from S-CGIB under the random splitting protocol. The results of HiMol are reproduced using the authors' publicly available code with the same 250K pretraining corpus and downstream splitting protocol to ensure a fair comparison. 

\subsection{Results}
Table \ref{tab:main_results} summarizes the performance comparison between MolCHG and 16 self-supervised pretraining baselines across nine molecular property prediction benchmarks. MolCHG achieves the best performance on 7 out of 9 datasets, covering both classification and regression tasks, and remains competitive on the remaining two.
\par

Among the classification tasks, MolCHG consistently outperforms all baselines on BACE, BBBP, ClinTox, and Tox21. Compared with the strongest baseline S-CGIB, MolCHG improves ROC-AUC by 0.74 percentage points on BACE ($86.46\to 87.20$), 0.71 on BBBP ($88.75\to 89.46$), 2.25 on ClinTox ($78.58\to 80.83$), and 1.57 on Tox21 ($80.94\to 82.51$). Notably, the improvement on ClinTox is particularly substantial, where MolCHG surpasses all prior methods by a clear margin. On SIDER and HIV, MolCHG achieves results comparable to the best-performing baselines (GraphFP and S-CGIB, respectively), with differences within one standard deviation.
\par

On all three regression tasks, MolCHG achieves the lowest RMSE. Compared with the previous best results, MolCHG reduces RMSE from 0.816 to 0.797 on ESOL, from 1.648 to 1.523 on FreeSolv, and from 0.762 to 0.745 on Lipophilicity. The improvement on FreeSolv is especially notable (a 7.6\% relative reduction), suggesting that the multi-level representations learned by MolCHG are particularly effective for tasks where both local functional group characteristics and global molecular topology contribute to the target property.
\par

Compared with node-level pretraining methods (ContextPred, AttrMasking, EdgePred), MolCHG achieves substantial improvements across all datasets, confirming the advantage of learning representations at multiple structural granularities rather than focusing solely on atom-level objectives. Compared with contrastive learning methods (Infomax, JOAO, JOAOv2, GraphCL, GraphLoG), MolCHG also demonstrates consistent superiority, indicating that the proposed level-specific predictive objectives provide more targeted supervision signals than general-purpose contrastive augmentation strategies. Among fragment-level pretraining methods, HiMol is the most closely related to our approach as it also constructs a hierarchical molecular graph. MolCHG outperforms HiMol on all 9 datasets, with notable improvements on BBBP (89.46\% vs 82.24\%) and Tox21 (82.51\% vs 80.16\%). These gains can be attributed to two key differences: the incorporation of bond-level information as an independent structural layer, and the design of level-specific pretraining objectives that provide complementary supervision across the hierarchy.
\section{Ablation Study}

\subsection{Ablation on Graph Structures}

\begin{table*}[t]
\centering
\caption{Ablation study on graph structures. Classification tasks are evaluated by ROC-AUC ($\uparrow$) and the regression task by RMSE ($\downarrow$).}
\label{tab:ablation_graph}
\begin{tabular}{lcccccc}
\hline
Graph Structure & BACE$\uparrow$ & BBBP$\uparrow$ & ClinTox$\uparrow$ &Tox21$\uparrow$ & HIV$\uparrow$&Lipophilicity$\downarrow$ \\
\hline
Atom Graph & $71.61\pm1.41$ & $74.96\pm1.43$ & $67.33\pm0.83$ & $69.54\pm 0.56$ &$68.29\pm 0.79$ & $1.137\pm0.027$ \\
Hierarchical Graph & $74.34\pm0.68$ & $76.59\pm0.72$ & $68.52\pm0.94$  &$70.81\pm0.71$ & $70.16\pm0.59$ & $1.073\pm0.013$ \\
\rowcolor{gray!15}
Compositional Hierarchical Graph & \boldmath{$76.33\pm0.80$} & \boldmath{$77.97\pm0.95$} & \boldmath{$69.97\pm1.45$} &\boldmath{$72.10\pm0.67$} &\boldmath{$71.42\pm0.80$} &  \boldmath{$1.057\pm0.016$} \\
\hline
\end{tabular}
\end{table*}

\begin{table*}[htbp]
\centering
\caption{Sensitivity analysis of pretraining loss weights. Classification tasks are evaluated by ROC-AUC (\%) ($\uparrow$) and the regression task by RMSE ($\downarrow$). The default setting is $\lambda_{ab}=0.2$, $\lambda_{frag}=0.4$, $\lambda_{topo}=0.4$, $\lambda_{scaf}=0.4$.}
\label{tab:sensitivity}
\footnotesize
\begin{tabular}{cccc|cccccc}
\toprule
$\lambda_{ab}$ & $\lambda_{frag}$ & $\lambda_{topo}$ & $\lambda_{scaf}$ & BACE $\uparrow$ & BBBP $\uparrow$ & ClinTox $\uparrow$ & Tox21 $\uparrow$ & HIV $\uparrow$ & Lipophilicity $\downarrow$ \\
\midrule
\rowcolor{gray!15}
0.2 & 0.4 & 0.4 & 0.4 &\boldmath{$87.20\pm0.86$} & \boldmath{$89.46\pm0.54$} & \boldmath{$80.83\pm1.69$} & \boldmath{$82.51\pm1.20$} & \boldmath{$77.83\pm0.67$} & \boldmath{$0.745\pm0.036$} \\
\midrule
0.1 & 0.4 & 0.4 & 0.4 &$86.41\pm1.04$  &$88.63\pm1.28$  &$80.18\pm1.31$ &$81.28\pm1.12$  &$76.54\pm1.02$  &  $0.761\pm0.030$ \\
0.4 & 0.4 & 0.4 & 0.4 &$86.29\pm1.07$  &$88.39\pm1.21$
&$79.71\pm1.48$ &$81.79\pm0.77$  &$76.97\pm1.01$  &  $0.770\pm0.029$\\
\midrule
0.2 & 0.2 & 0.4 & 0.4 &$86.66\pm0.77$  &$87.09\pm1.39$  &$80.46\pm1.57$  &$82.18\pm1.01$ &$77.22\pm0.57$  &  $0.756\pm0.026$\\
0.2 & 0.6 & 0.4 & 0.4 &$86.25\pm0.60$  &$88.16\pm1.56$  &$80.26\pm1.59$  &$81.60\pm0.92$ &$77.12\pm0.75$  &  $0.759\pm0.026$\\
\midrule
0.2 & 0.4 & 0.2 & 0.2 &$86.96\pm0.94$  &$88.43\pm1.19$  &$80.54\pm1.60$  &$81.41\pm0.89$ &$77.06\pm0.90$ & $0.758\pm0.015$\\
0.2 & 0.4 & 0.6 & 0.6 &$86.47\pm0.81$  &$87.56\pm0.71$  &$80.48\pm1.62$  &$81.45\pm1.42$ &$76.50\pm0.76$ & $0.767\pm0.029$\\
\bottomrule
\end{tabular}
\end{table*}

To validate the contribution of the proposed Compositional Hierarchical Graph, we compare three graph structures under identical training conditions: (1) the standard atom graph, (2) the hierarchical graph that organizes atoms, fragments, and a virtual graph node into a multi-level structure but without an independent bond graph, and (3) our Compositional Hierarchical Graph that incorporates the bond graph as an independent structural layer. All three variants are trained from random initialization on a 5-layer GIN without pretraining, so that the observed differences can be attributed solely to the graph structure. We evaluate on six downstream datasets: BACE, BBBP, ClinTox, HIV, Tox21 and Lipophilicity. 
\par

As shown in Table~\ref{tab:ablation_graph}, introducing the hierarchical structure with fragment and graph nodes consistently improves performance over the atom graph alone, confirming the benefit of multi-level organization. More importantly, further incorporating the bond graph into the hierarchy yields additional gains across all six datasets. For instance, the Compositional Hierarchical Graph improves ROC-AUC over the hierarchical graph by 2.0 percentage points on BACE ($74.34\to 76.33$), 1.4 on BBBP ($76.59\to 77.97$), and 1.5 on ClinTox ($68.52\to 69.97$), while reducing RMSE on Lipophilicity from 1.073 to 1.057. Similar improvements are observed on Tox21 ($70.81\to 72.10$) and HIV ($70.16\to 71.42$). These results demonstrate that elevating bond-level information from auxiliary edge attributes to independent node representations provides the model with richer structural semantics, validating the core design principle of our proposed graph structure.
\par

\subsection{Ablation on Pretraining Objectives}
To assess the contribution of each pretraining objective, we systematically remove individual loss terms and groups of losses from the full framework. As shown in Table~\ref{tab:ablation_loss}, all pretraining variants substantially outperform the No Pretraining baseline, confirming the overall effectiveness of self-supervised pretraining on the Compositional Hierarchical Graph. Among the single-loss ablations, removing any individual objective leads to a consistent performance degradation across all six datasets. Notably, removing $L_{frag}$ causes the largest average drop among classification tasks (e.g., BACE decreases from 87.20\% to 84.14\%, Tox21 from 82.51\% to 78.55\%), highlighting the importance of injecting domain-relevant functional group knowledge into the intermediate layer. Removing $L_{topo}$ results in a pronounced decline on HIV (from 77.83\% to 75.34\%), suggesting that topological fingerprint prediction is particularly beneficial for tasks involving large-scale molecular screening where global connectivity patterns are informative.
\par

More importantly, removing an entire level of objectives produces substantially larger performance drops than removing any single loss. The w/o Graph-level variant, which retains only atom-bond and fragment-level supervision, shows notable degradation on BBBP ($89.46\% \to 84.30\%$) and HIV ($77.83\% \to 74.38\%$). Similarly, the w/o Atom-bond \& Fragment-level variant, which retains only graph-level supervision, yields the lowest performance among all pretraining variants on most datasets (e.g., ClinTox drops to 76.03\%, Lipophilicity increases to 0.814). The fact that both level-removal variants perform substantially worse than any single-loss variant demonstrates that the supervision signals from different levels are complementary rather than redundant: graph-level objectives alone cannot compensate for the absence of the local compositional and fragment-level semantics, and vice versa. These results validate the core design principle of MolCHG---that multi-level pretraining objectives operating across the full hierarchy are essential for learning comprehensive molecular representations.

\begin{table*}[htbp]
\centering
\caption{Ablation study on pretraining objectives. Classification tasks are evaluated by ROC-AUC (\%) ($\uparrow$) and the regression task by RMSE ($\downarrow$).}
\label{tab:ablation_loss}
\footnotesize
\begin{tabular}{l|cccccc}
\toprule
Variant & BACE $\uparrow$ & BBBP $\uparrow$ & ClinTox $\uparrow$ & Tox21 $\uparrow$ & HIV $\uparrow$ & Lipophilicity $\downarrow$ \\
\midrule
No Pretraining & $76.33\pm0.80$ & $77.97\pm0.95$ & $69.97\pm1.45$ & $72.10\pm0.67$ & $71.42\pm0.80$ & $1.057\pm0.016$ \\
\midrule
w/o $\mathcal{L}_{ab}$ &$85.53\pm1.08$ &$86.23\pm1.47$  &$77.60\pm1.26$  &$79.62\pm1.02$  &$76.93\pm1.00$ & $0.789\pm0.034$ \\
w/o $\mathcal{L}_{frag}$ &$84.14\pm1.53$  &$87.08\pm2.29$  &$77.20\pm2.44$  &$78.55\pm0.68$  &$76.20\pm1.70$  & $0.797\pm0.035$ \\
w/o $\mathcal{L}_{topo}$ &$85.93\pm0.99$  &$87.94\pm1.32$  &$78.65\pm1.71$  &$78.37\pm0.92$  &$75.34\pm0.58$  &$0.788\pm0.024$  \\
w/o $\mathcal{L}_{scaf}$ &$84.60\pm1.64$  &$87.62\pm0.80$  &$77.68\pm1.36$  &$78.99\pm0.99$  &$76.62\pm0.78$  &$0.767\pm0.025$  \\
\midrule
w/o Graph-level &$83.69\pm1.44$  &$84.30\pm1.56$  &$76.87\pm1.88$  &$77.54\pm0.91$  &$74.38\pm1.78$  &$0.809\pm0.025$  \\
w/o Atom-bond \& Fragment-level &$82.83\pm1.17$  &$85.51\pm0.92$  &$76.03\pm2.60$  &$76.70\pm1.06$  &$74.09\pm1.80$  &$0.814\pm0.026$  \\
\midrule
\rowcolor{gray!15}
MolCHG (Full) & \boldmath{$87.20\pm0.86$} & \boldmath{$89.46\pm0.54$} &\boldmath{$80.83\pm1.69$}  &\boldmath{$82.51\pm1.20$}  & \boldmath{$77.83\pm0.67$} & \boldmath{$0.745\pm0.036$} \\
\bottomrule
\end{tabular}
\end{table*}

\subsection{Sensitivity Analysis of Loss Weights}

We examine the robustness of MolCHG with respect to the pretraining loss weights by varying each weight while keeping others at their default values ($\lambda_{ab}=0.2,\lambda_{frag}=0.4, \lambda_{topo}=0.4,\lambda_{scaf}=0.4$). As shown in Table~\ref{tab:sensitivity}, MolCHG maintains stable performance across all weight configurations, with no variant causing drastic performance degradation. This indicates that the framework is robust to the specific choice of loss weights within a reasonable range.
\par

Among the individual weights, $\lambda_{ab}$ is the most sensitive to changes: increasing it to 0.4 leads to a consistent drop across most datasets (e.g., BBBP from $89.46\%$ to $88.39\%$, Lipophilicity from $0.745$ to $0.770$), confirming our design choice of assigning a smaller weight to the contrastive loss due to its inherently larger magnitude. For $\lambda_{frag}$, reducing it to 0.2 causes a notable decline on BBBP ($89.46\%\to 87.09\%$), suggesting that sufficient weight on the functional group prediction task is important for injecting domain-relevant semantics into the fragment-level representations. For the graph-level weights varied jointly, over-emphasizing them to $0.6$ results in more pronounced degradation than reducing them to $0.2$ (e.g., HIV drops from $77.83\%$ to $76.50\%$), indicating that excessively strong graph-level supervision may overshadow the fine-grained local patterns captured by the lower-level objectives. Overall, the default configuration achieves the best balance among the multi-level objectives.

\begin{figure*}[t]
\centering
\includegraphics[width=\textwidth]{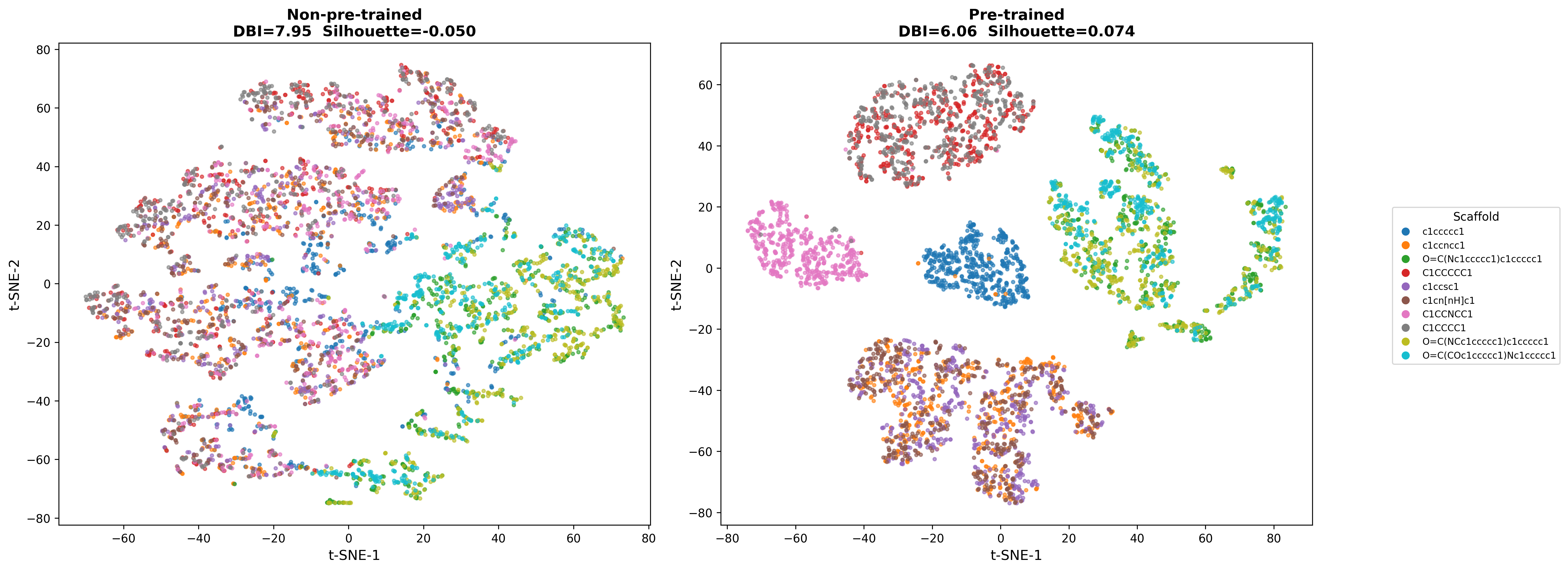}
\caption{t-SNE visualization of graph-level representations colored by Murcko scaffolds. Left: Randomly initialized model. Right: pre-trained model.}
\label{fig:Tsne1}
\end{figure*}

\section{Visualization}

To examine whether each level of the proposed multi-level pretraining framework fulfills its intended role, we conduct t-SNE \cite{van2008visualizing} visualizations of the learned representations at the graph, fragment, and bond levels. The clustering quality is quantified by DBI \cite{davies1979cluster} and Silhouette Score \cite{rousseeuw1987silhouettes} computed on the original 300-dimensional representation space rather than on the t-SNE projections, since t-SNE is a nonlinear dimensionality reduction technique that can distort inter-cluster distances. We expect pretraining to improve clustering metrics across all levels, indicating that the learned representations capture semantically meaningful structure. However, perfect clustering by any single categorical label is neither expected nor desirable: a representation that separates one label perfectly would have collapsed its information content to that label alone, sacrificing the multi-faceted chemical semantics needed for diverse downstream tasks. The moderate inter-cluster overlap observed at each level therefore reflects this intended property rather than a deficiency.

\subsection{Graph-level representation analysis}
To verify whether the graph-level pretraining objectives enable the graph node to capture the global topological organization of molecules, we visualize the graph node representations using t-SNE \cite{van2008visualizing}. We select the 10 most frequent Murcko scaffolds \cite{bemis1996properties} from the pretraining corpus and sample 500 molecules per scaffold. As shown in Figure \ref{fig:Tsne1}, the randomly initialized model produces interleaved representations with no meaningful clustering structure (DBI = $7.95$, Silhouette = $-0.050$). After pretraining, molecules sharing the same scaffold are organized into more distinguishable regions (DBI = $6.06$, Silhouette = $0.074$), and structurally similar scaffolds such as benzene and pyridine form neighboring clusters, indicating that the learned representations preserve underlying structural relationships.
The moderate inter-cluster overlap is consistent with our expectation, as molecules sharing the same scaffold can differ substantially in side chains and substituents, and the graph-level representation encodes the full molecular topology beyond the scaffold alone.

\begin{figure*}[t]
\centering
\includegraphics[width=\textwidth]{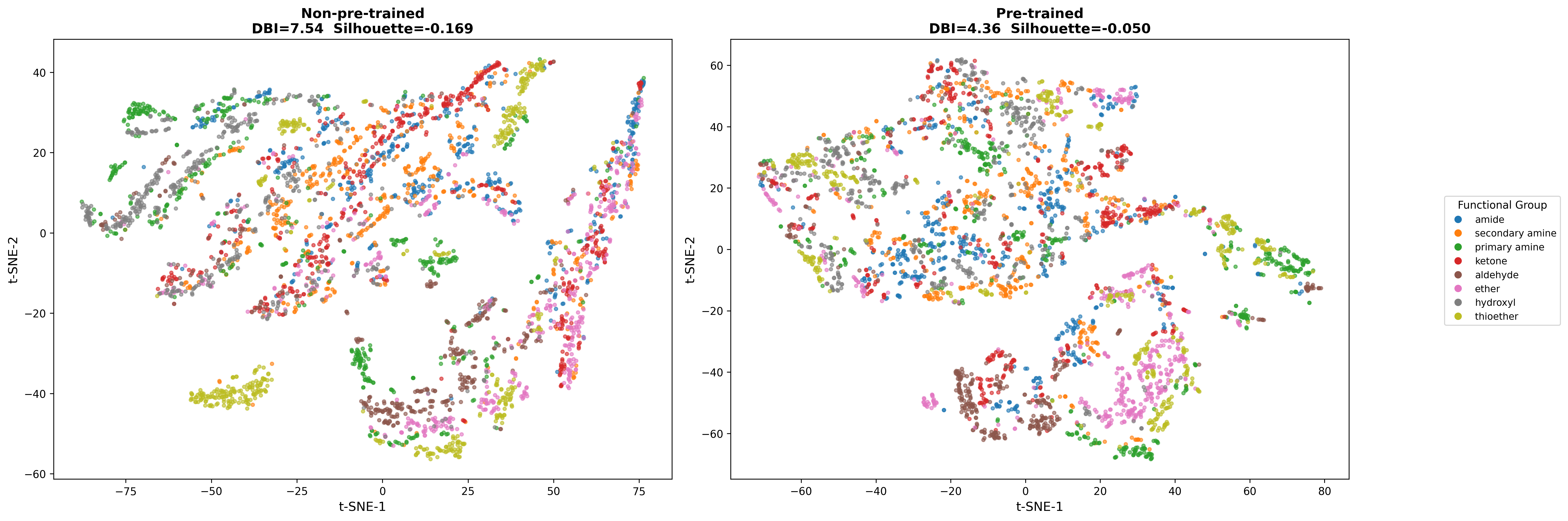}
\caption{t-SNE visualization of fragment-level representations colored by functional group types. Left: randomly initialized model. Right: pre-trained model.}
\label{fig:Tsne2}
\end{figure*}

\begin{figure*}[t]
\centering
\includegraphics[width=\textwidth]{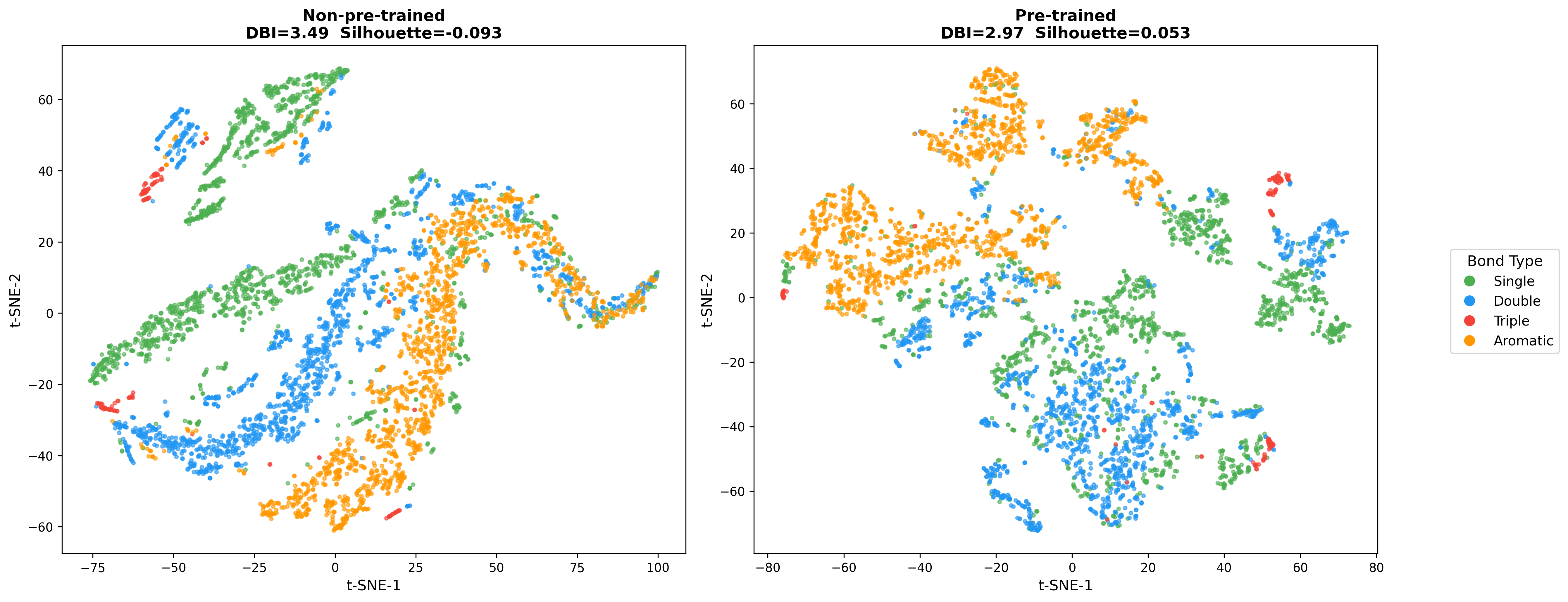}
\caption{t-SNE visualization of bond-level representations colored by bond types. Left: randomly initialized model. Right: pretrained model.}
\label{fig:Tsne3}
\end{figure*}

\subsection{Fragment-level representation analysis}
The fragment-level pretraining objective is designed to inject domain-relevant semantic knowledge into the intermediate layer of the hierarchy by predicting the presence of functional groups within each fragment. We select 8 common functional groups and sample 500 fragments per group for t-SNE visualization. Figure \ref{fig:Tsne2} reveals that pretraining substantially improves the separability of fragment representations across functional group categories (DBI: $7.54 \to 4.36$, Silhouette: $-0.169 \to -0.050$), with groups such as aldehyde, ether, and primary amine forming more distinguishable regions. The residual overlap reflects the context-dependent nature of fragment representations, as the same functional group can exhibit different chemical behavior depending on its molecular environment.

\subsection{Bond-level representation analysis}

In conventional molecular graphs, bond information is encoded as edge attributes that are consumed during message construction but never updated as independent entities. Our Compositional Hierarchical Graph addresses this limitation by organizing bonds as an independent node layer. We visualize bond node representations using t-SNE, colored by bond type (single, double, triple, and aromatic), with 1,500 bonds per type sampled from 2,000 molecules. As shown in Figure \ref{fig:Tsne3}, the randomly initialized model already exhibits partial separation because the initial bond node features encode bond-type descriptors (DBI = $3.49$, Silhouette = $-0.093$). After pretraining, separation becomes substantially clearer (DBI = $2.97$, Silhouette = $0.053$). The improvement over the random baseline thus reflects how pretraining refines and contextualizes the initial type descriptors through message passing, rather than discovering bond-type information from scratch. Aromatic and triple bonds form more coherent clusters, while the remaining overlap among single and double bonds reflects that bond type is a coarse label that does not fully determine chemical behavior, and the atom-bond cross-view contrastive objective encourages bond nodes to capture contextual semantics beyond their initial type descriptors.

\section{Conclusion}
In this work, we proposed MolCHG, a multi-level self-supervised pretraining framework built upon the Compositional Hierarchical Graph, which introduces a bond graph as an independent structural layer and enables fragment nodes to aggregate atom-level and bond-level semantics on an equal footing. Three level-specific pretraining objectives—--atom–bond cross-view contrastive learning, fragment-level functional group prediction, and graph-level structure prediction—--provide complementary supervision signals spanning from local compositional patterns to global molecular topology. Extensive experiments on nine MoleculeNet benchmarks demonstrated that MolCHG achieved the best performance on seven datasets and remained competitive on the remaining two. Ablation studies confirmed that each component contributed meaningfully to the overall performance and that multi-level supervision was essential for learning comprehensive molecular representations.

\section{Key points}
\begin{itemize}
    \item We propose the Compositional Hierarchical Graph, a heterogeneous molecular graph with four co-existing node types across three semantic levels, in which fragment nodes are explicitly connected to both atom nodes and bond nodes from two parallel structural layers. This design elevates bond information from auxiliary edge attributes to independently evolving node representations and enables atom-level and bond-level semantics to be aggregated at the fragment level on an equal footing.
    \item We design three level-specific self-supervised pretraining objectives that jointly operate over the Compositional Hierarchical Graph: an atom–bond cross-view contrastive task aligning the atom-view and bond-view representations of each fragment, a fragment-level functional group prediction task injecting domain-relevant chemical knowledge, and graph-level structure prediction tasks encoding global molecular topology.
    \item MolCHG achieves the best performance on seven out of nine MoleculeNet benchmarks across both classification and regression tasks, and remains competitive on the rest. Ablation studies and representation visualizations confirm that the supervision signals from different levels are complementary and that each component contributes meaningfully to the overall performance.
\end{itemize}

\section{Competing interests}
No competing interest is declared.

\section{Author contributions statement}
X.L. conceived the study, designed the methodology, implemented the framework, conducted the experiments, and wrote the original manuscript. Z.L. contributed to the data preprocessing and assisted with the experimental evaluation. H.L. provided supervision and contributed to the revision of the manuscript. All authors reviewed and approved the final manuscript.

\section{Data availability}
The dataset and code are available at \url{https://github.com/lhb0189/MolCHG}
\section{Acknowledgments}
This work was supported by National Natural Science Foundation of China (Grant: 11101071). The authors also acknowledge the Center for High Performance Computing, University of Electronic Science and Technology of China, for computational support."

\bibliographystyle{unsrt}
\bibliography{reference}

\end{document}